\documentclass[runningheads]{llncs}
\usepackage[T1]{fontenc}
\usepackage{cite}
\usepackage{amsmath,amssymb,amsfonts}
\usepackage{graphicx}
\usepackage{textcomp}
\usepackage{booktabs}
\usepackage{multirow}
\usepackage[colorlinks=false,breaklinks=true]{hyperref}
\usepackage{xurl}

\begin{document}

\title{Hierarchical Curriculum Learning for Surgical Video Generation: Focusing on Bilateral Axillo-Breast Approach Robotic Thyroidectomy Datasets}
\titlerunning{Hierarchical Curriculum Learning for Surgical Video Generation}

\author{Youngbin Kong\inst{1,2} \and
Seungho Lee\inst{5} \and
Jae Kyung Shin\inst{5} \and
Jong-hyuk Ahn\inst{6} \and
Hyoun-Joong Kong\inst{2,3,4,\star} \and
Kyu Eun Lee\inst{5,7,\star}}
\authorrunning{Y. Kong et al.}

\institute{Interdisciplinary Program in Medical Informatics, Seoul National University College of Medicine, Seoul, Republic of Korea
\and
Department of Transdisciplinary Medicine, Seoul National University Hospital, Seoul, Republic of Korea
\and
Institute of Convergence Medicine with Innovative Technology, Seoul National University Hospital, Seoul, Republic of Korea
\and
Department of Medicine, Seoul National University College of Medicine, Seoul, Republic of Korea
\and
Department of Surgery, Seoul National University Hospital, Seoul, Republic of Korea
\and
YONSEI WITH-YOU Surgery Clinic, Seoul, Republic of Korea
\and
Seoul National University College of Medicine, Seoul, Republic of Korea\\[4pt]
$^{\star}$~Corresponding authors}

\maketitle

\begin{abstract}
High-fidelity surgical video generation is increasingly important for surgical education, simulation, and downstream AI training, as conventional simulation pipelines are limited by cost, realism, and scenario diversity. Adapting generic text-to-video diffusion models to surgical procedures remains difficult, however, due to complex anatomical structures, intricate tool--tissue interactions, the inherently hierarchical nature of surgical workflows, and the scarcity of curated training data outside a few well-studied procedures. We propose a surgical video generation framework that combines hierarchical procedural annotation with curriculum transfer learning. Surgical procedures are decomposed into three levels (phase, task, and action) and a pretrained text-to-video diffusion model is progressively adapted from coarse to fine-grained surgical context, supported by a dynamic sampling strategy that handles variable-resolution and variable-length surgical clips. We instantiate the framework on Bilateral Axillo-Breast Approach robotic thyroidectomy, a procedure for which no curated benchmark currently exists. Relative to the preceding task-level model, the final action-level model reduces FVD by 20.1\% and FID by 15.7\%, and a matched-budget ablation indicates that these gains arise from the curriculum design rather than from additional training. A user study with six clinicians rates visual realism highly while highlighting remaining gaps in anatomical accuracy and clinical reliability. The proposed framework offers a step toward AI-driven surgical simulation in procedures where curated data are scarce.
\end{abstract}

\section{Introduction}
\label{sec:introduction}
Surgical simulation and computer-assisted training systems have become integral components of modern surgical education, allowing trainees to acquire technical skills in a risk-free, standardized, and objectively assessable environment, and decoupling the resident learning curve from patient risk~\cite{ref12_cardoso2023exploring,ref13_demontbrun2012simulation}. The field has progressed from low-fidelity box trainers and cadaveric or animal models~\cite{ref15_sharma2020lowcost} to virtual reality (VR) systems that enable repeatable practice without spatiotemporal constraints, with growing evidence of its effectiveness in surgical training~\cite{ref16_peng2024virtual,ref17_rogister2020virtual,ref18_lucas2008training}. More recently, real surgical videos have emerged as a richer simulation source~\cite{ref19_cho2024surgen}, motivated by evidence that video-based curricula improve resident preparation and intraoperative participation~\cite{ref20_kim2022videobased}, and computer-assisted intervention approaches further combine deep models with surgical video to enable 3D reconstruction and immersive simulation~\cite{ref21_liu2024endogaussian,ref22_yang2024deform3dgs,ref23_wang2024realistic}. Yet these approaches remain limited by the size and diversity of public surgical video datasets, which makes large-scale, scenario-diverse simulation difficult to extend across surgical disciplines.

Generative artificial intelligence has recently enabled high-fidelity synthesis across text, image, and video modalities~\cite{ref39_ho2020denoising,ref46_openai2024sora}, with diffusion-based video models in particular emerging as the dominant paradigm due to their training stability, robustness to mode collapse, and high sample quality~\cite{ref38_gainetdinov2023diffusion,ref39_ho2020denoising}. Modern text-to-video systems such as Sora~\cite{ref46_openai2024sora} and Open-Sora~\cite{ref47_zheng2024opensora} achieve high visual realism and temporal coherence on general-domain content, raising the prospect of AI-generated surgical videos as scalable training material, augmentation for downstream surgical AI, and a substrate for surgical planning. However, applying these generic video generation models directly to the surgical domain remains challenging. Surgical scenes contain complex anatomical structures, subtle tissue deformations, and intricate tool--tissue interactions, and surgical procedures inherently follow a hierarchical structure progressing from operational phases through intermediate tasks down to specific instrument actions. Generic text-to-video models often fail to capture this nuanced hierarchy, leading to temporal inconsistencies, physically implausible tool motion, and anatomical inaccuracies. Moreover, existing surgical video generation work has concentrated on procedures for which curated public datasets such as CholecT50~\cite{ref62_nwoye2022rendezvous} or Kvasir-Capsule~\cite{ref63_smedsrud2021kvasir} already exist, leaving procedures with distinct anatomical access or instrumentation largely unaddressed.

We focus on Bilateral Axillo-Breast Approach (BABA) robotic thyroidectomy as a representative example of such procedures. Thyroidectomy is among the most common endocrine surgeries worldwide, and BABA, a remote-access technique developed at Seoul National University Hospital~\cite{ref1_choe2007baba} and later adapted to robotic surgery, provides an anatomically symmetric midline view that closely resembles the open surgical perspective~\cite{ref11_lee2011bilateral}. BABA combines an abundance of high-quality intraoperative video with the absence of a curated public dataset, leaving it unexplored in recent surgical video generation research. This combination of rich raw footage and missing curated benchmarks makes BABA an ideal testbed for data-efficient adaptation of large pretrained video models to a new procedural domain.

Recent advances in surgical workflow analysis suggest a path forward. Surgical procedures are inherently hierarchical, ranging from the entire operation down to phases, tasks, and actions, and recent works exploit this structure for surgical video understanding~\cite{ref64_yuan2024hecvl,ref65_yuan2024procedure}. In parallel, curriculum learning, inspired by human learning, orders training data by difficulty and has been explored in medical image analysis~\cite{ref66_liu2022acpl,ref67_singh2023see}. However, while these advances enable richer surgical understanding, they have been applied almost exclusively to surgical \emph{analysis}, with very limited exploration of how hierarchical procedural structure can be reflected in surgical video \emph{generation}: the gap addressed by this paper.

To this end, we introduce a surgical video generation framework based on \textbf{hierarchical annotation and curriculum transfer learning}, with BABA robotic thyroidectomy as the target procedure. We construct a Phase--Task--Action three-level annotation of clinical BABA videos, and on top of the Open-Sora~v1.2 architecture~\cite{ref47_zheng2024opensora}, which combines a T5-XXL text encoder~\cite{ref68_raffel2020exploring}, a 3D variational autoencoder (VAE), and a Spatial-Temporal Diffusion Transformer (STDiT), we fine-tune only STDiT through three progressive stages aligned with the annotation hierarchy. A dynamic bucket-based sampling strategy further enables efficient training on variable-resolution and variable-length surgical clips.

To our knowledge, this is the first work to incorporate explicit hierarchical procedural structure into surgical video \emph{generation} rather than recognition. Our contributions are summarized as follows:
\begin{enumerate}
    \item \textbf{Hierarchical curriculum transfer learning.} We propose a three-stage strategy that progressively adapts a pretrained video diffusion backbone from coarse Phase-level to fine-grained Action-level supervision, mirroring the surgical workflow hierarchy while keeping the text encoder and VAE frozen. To align data difficulty with each stage, we repurpose Open-Sora's dynamic bucket sampling~\cite{ref47_zheng2024opensora} as a stage-conditioned curriculum control rather than treating it as a new mechanism.
    \item \textbf{Phase--Task--Action annotation for generation.} We define and apply a three-level temporal annotation schema for BABA robotic thyroidectomy, providing structured text supervision tailored to generative, not discriminative, training.
    \item \textbf{Data-efficient adaptation to a benchmark-free procedure.} We instantiate the framework on BABA, for which no curated public dataset exists, and validate it through quantitative metrics, an ablation that isolates the curriculum effect from additional training, and a clinical user study with six clinicians.
\end{enumerate}

\section{Related Work}
\label{sec:related_works}

\subsection{General-Domain Video Generation}
Video generation has progressed rapidly with the rise of diffusion models~\cite{ref40_ho2022video}, after early efforts dominated by GAN-based approaches~\cite{ref41_melnik2024video}. Video Diffusion Models (VDM)~\cite{ref40_ho2022video} first applied diffusion to video generation by extending the 2D U-Net used in image diffusion to a 3D U-Net, enabling unconditional and conditional generation while improving inter-frame consistency. Make-A-Video~\cite{ref42_singer2022makeavideo} introduced a strategy that augments a pretrained text-to-image model with additional spatial and temporal layers, enabling video synthesis without paired text--video data. Imagen Video~\cite{ref43_ho2022imagen} adopted a cascaded diffusion pipeline that first generates low-resolution videos and then performs multi-stage spatial and temporal super-resolution to reach high-definition output.

More recent models emphasize temporal consistency and high-resolution generation. ART-V~\cite{ref44_weng2024artv} introduces masked diffusion to enable autoregressive frame generation while maintaining temporal coherence over long sequences. Lumiere~\cite{ref45_bartal2024lumiere} uses a Space-Time U-Net (STUNet) to generate all frames jointly, replacing the keyframe-then-interpolate paradigm with a holistic low-resolution generation followed by super-resolution. OpenAI's Sora~\cite{ref46_openai2024sora} produces up to 60-second high-resolution ($1920\times1080$) videos with complex multi-agent scenes, while Open-Sora~\cite{ref47_zheng2024opensora} provides an open-source counterpart based on a Spatial-Temporal Diffusion Transformer (STDiT) and a 3D VAE, supporting up to 16-second videos at up to 720p with variable aspect ratios.

While these models achieve high realism on general-domain content, they do not encode the conceptual structure or domain-specific data required for highly structured procedural video, such as surgery, and thus cannot be applied directly to surgical video generation.

\subsection{Surgical Video Generation in the Medical Domain}
Surgical video generation poses unique challenges beyond those of general-domain video~\cite{ref48_yeganeh2024visage}, requiring high accuracy and realism for use in medical education and simulation~\cite{ref49_mcguire2018future}. Early work applied neural style transfer~\cite{ref50_gatys2016image}, which combines content-feature reconstruction with texture synthesis, to enhance the realism of simulation-rendered images. SurReal~\cite{ref51_luengo2018surreal} transferred the style of real surgical videos onto simulation renderings, preserving structural correctness while greatly improving visual realism. Real-time style transfer algorithms have also been proposed to improve the realism of laparoscopic surgical simulators while maintaining temporal smoothness~\cite{ref52_shao2022realtime}, applying the appearance of real surgical images to silicone-based phantoms.

A complementary line of work focuses on 3D mesh stylization and texture synthesis. Paget \textit{et al.}~\cite{ref54_paget2005framework} proposed a coherent texturing framework that generates tileable variable textures from intraoperative images and maps them onto 3D mesh geometry. More recently, MeshBrush~\cite{ref55_han2024meshbrush} performs neural mesh stylization that preserves temporal consistency for endoscopic video by learning vertex-wise textures on meshes derived from patient imaging.

Beyond texturing and stylization, an emerging direction generates surgical videos directly. Endora~\cite{ref56_li2024endora} is among the first surgical video generation frameworks targeting endoscopic simulation, combining a latent diffusion model with a spatial-temporal video transformer and a DINO~\cite{ref57_caron2021emerging} feature prior that enhances inter-frame consistency.
SurGen~\cite{ref19_cho2024surgen} is a text-guided diffusion model for surgical video synthesis built on CogVideoX~\cite{yang2025cogvideox}, generating 480p clips of 49 frames. SurgSora~\cite{ref31_chen2024surgsora} extends this direction with a more controllable design that uses a single input frame and motion cues, comprising a Dual Semantic Injector, a Decoupled Flow Mapper, and a Trajectory Controller, and decoupling RGB-D features from optical flow. VISAGE~\cite{ref48_yeganeh2024visage} synthesizes laparoscopic surgical videos via action scene graphs whose nodes represent instruments and organs and whose edges represent the performed actions, capturing the spatial and relational dynamics of surgical activity through diffusion models. MedSora~\cite{ref59_wang2024medsora} integrates an attention-Mamba video diffusion backbone~\cite{ref60_gu2023mamba}, an optical-flow-based representation alignment module that enhances pixel-level inter-frame attention, and a frequency-compensated video VAE designed to reduce information loss for medical features. In ophthalmic surgery, Ophora~\cite{ref61_li2025ophora} demonstrates that natural-language-guided ophthalmic surgical videos can be generated using a large dataset of more than 160K video--instruction pairs and a progressive video instruction tuning scheme that transfers spatiotemporal knowledge from general-domain pretraining.

These advances have, however, concentrated on surgical domains for which curated datasets such as CholecT50~\cite{ref62_nwoye2022rendezvous} and Kvasir-Capsule~\cite{ref63_smedsrud2021kvasir} already exist, leading to strong domain dependency and limited generalization. Procedures with distinct anatomical access and instrumentation, such as BABA robotic thyroidectomy, cannot directly benefit from these methods.

\subsection{Hierarchical Annotation and Curriculum Learning}
Surgical procedures are inherently hierarchical, ranging from the entire operation down to phases, tasks, and actions. Recent works exploit this structure for surgical workflow analysis. HecVL~\cite{ref64_yuan2024hecvl} and PeskaVLP~\cite{ref65_yuan2024procedure} pair short clips with atomic action descriptions, longer segments with phase summaries, and full videos with overall summaries, enabling models to capture both short-range and long-range dependencies and yielding strong representations for downstream tasks such as zero-shot phase recognition, with PeskaVLP further using LLMs to refine and augment these hierarchical texts.

In parallel, curriculum learning has been explored in medical image analysis through progressive easy-to-hard schedules~\cite{ref67_singh2023see} and anti-curriculum selection of informative samples~\cite{ref66_liu2022acpl}. While these advances enable richer surgical understanding, they operate in a discriminative setting in which hierarchy guides representation learning for recognition rather than constraining a generative process. How hierarchical procedural structure should instead shape surgical video \emph{generation}, where the model must synthesize, not classify, phase-, task-, and action-level content, remains largely unexplored.

\section{Methodology}
\label{sec:methodology}
This section details the proposed framework for surgical video generation. We first describe the construction of the hierarchically annotated clinical dataset, then introduce the network architecture, and finally present the dynamic bucket-based sampling and three-stage curriculum transfer learning strategy.

\subsection{Hierarchically Annotated Surgical Video Dataset}

\subsubsection{Data Collection}
A total of 43 BABA robotic thyroidectomy videos were collected from Seoul National University Hospital. All operations were performed using the da~Vinci~Xi system, and each video covers the full procedure from start to end as a continuous recording at a resolution of $1920\times1080$. This study was approved by the Institutional Review Board (IRB) of Seoul National University Hospital (IRB No.~H-2207-162-1344). All patient identifiers were removed, and the data were stored and used in compliance with the institutional ethical guidelines. 

\subsubsection{Hierarchical Annotation}
\begin{figure}[htbp]
\centering
\includegraphics[width=\textwidth]{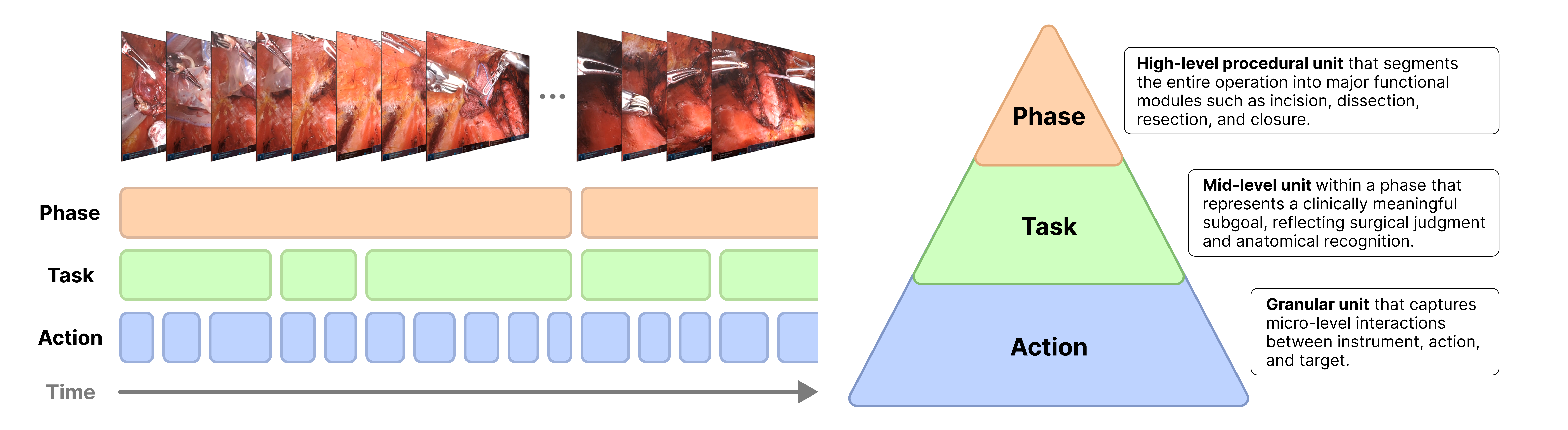}
\caption{Schematic and example of the hierarchical annotation system for surgical video. BABA robotic thyroidectomy is divided into six Phases (MID, LAT, INF, BER, SUP, END), each further decomposed into Tasks and Actions.}
\label{fig:hierarchical_annotation}
\end{figure}

To convert the structured surgical workflow into a form suitable for training, we adopt a Phase--Task--Action three-level Hierarchical Surgical Annotation Schema (Fig.~\ref{fig:hierarchical_annotation}). Phases are macro-level operational steps that modularize high-level surgical functions, tasks are intermediate clinical sub-goals reflecting surgeon judgment, instrument manipulation, and anatomical recognition, and actions are atomic micro-level units describing tool--motion--target interactions.

The Phase hierarchy is defined based on the six surgical modules of BABA robotic thyroidectomy proposed by Kim~\textit{et al.}~\cite{ref29_kim2024defining}: midline incision to isthmectomy (MID), lateral dissection (LAT), preservation of inferior parathyroid glands (INF), recurrent laryngeal nerve and Berry ligament dissection (BER), superior pole dissection (SUP), and specimen removal and closure (END). The Task hierarchy is derived through cognitive task analysis of intermediate procedural goals within each phase. The Action hierarchy follows the action triplet representation $\langle$\textit{instrument, verb, target}$\rangle$ proposed by Nwoye~\textit{et al.}~\cite{ref62_nwoye2022rendezvous}.

Annotations were performed by clinicians from the breast and endocrine surgery division of Seoul National University Hospital, who recorded start/end timestamps and textual captions at each level. As an exploratory methodological study, Task and Action annotations focused on the most generalizable End Phase (specimen removal and closure).

\subsubsection{Preprocessing and Dataset Construction}
Each video is segmented into clips along the Phase--Task--Action timestamps, then further split into fixed-length sequences of 120 frames at 24~FPS to balance training stability and surgical context. CSV-format metadata is generated for each sequence, mapping it to its caption and recording resolution, frame length, and aspect ratio for downstream sampling. Among the 43 cases, 4 were held out as a patient-level test set and the remaining 39 were used for training. At the Phase, Task, and Action levels, 20, 16, and 7 cases were annotated, respectively; after holding out the 4 test cases, 20, 12, and 7 cases remained for training. The per-level clip counts and training budgets are reported in Table~\ref{tab:training_schedule}.

\subsection{Network Architecture}
\begin{figure}[htbp]
\centering
\includegraphics[width=\textwidth]{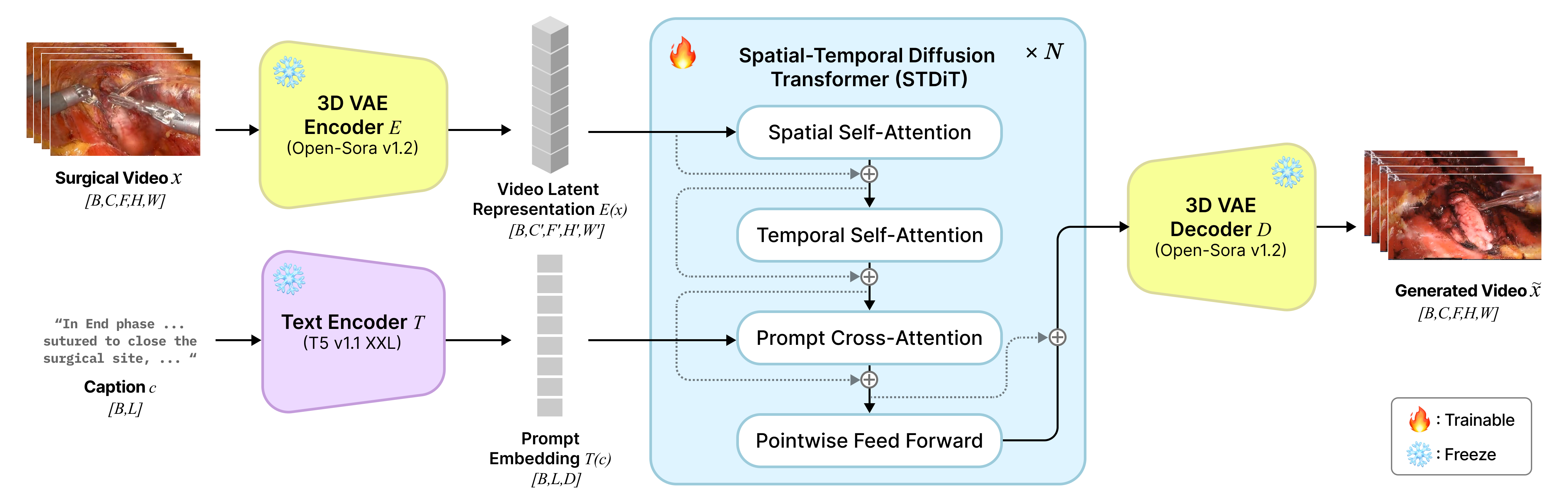}
\caption{Architecture of the surgical video generation model. The pipeline integrates a T5-XXL text encoder, the Open-Sora 3D VAE, and the Spatial-Temporal Diffusion Transformer (STDiT3).}
\label{fig:model_architecture}
\end{figure}

We adopt the Open-Sora~v1.2 architecture~\cite{ref47_zheng2024opensora}, an open-source Sora-like video generation model~\cite{ref46_openai2024sora}, pretrained on more than 20M training samples with approximately 35{,}000 NVIDIA~H100 GPU hours~\cite{opensora_v12_report}. The model supports up to 16-second clips at resolutions from 144p to 720p, and contains $\sim$1.1B parameters. The architecture comprises three modules (Fig.~\ref{fig:model_architecture}).

\textit{Text Encoder.} We use T5-XXL~\cite{ref68_raffel2020exploring} to encode surgical text prompts (e.g., \textit{``robotic instrument grasping tissue during lateral dissection''}) into rich contextual embeddings that condition STDiT.

\textit{3D VAE.} The Open-Sora v1.2 VAE compresses input videos into a low-dimensional spatiotemporal latent representation and decodes them back. It first compresses each frame spatially by a factor of 8 with a 2D VAE and then compresses the temporal dimension by a factor of 4 with a temporal VAE, producing more compact latents than a per-frame 2D VAE alone.

\textit{STDiT3.} The core diffusion model operates in the latent space and learns conditional generation between text embeddings and video latents. To remain computationally tractable, STDiT decouples spatial and temporal attention: each block first applies spatial attention within frames, then temporal attention across frames, achieving efficient yet expressive spatiotemporal modeling. We adopt the STDiT3-XL/2 configuration: $28$ blocks, hidden size $1152$, $16$ attention heads, patch size $(1,2,2)$, and approximately $1.12$B parameters.

\subsubsection{Preliminaries: Latent Rectified-Flow Generation}
Following Open-Sora~\cite{ref47_zheng2024opensora}, the frozen 3D VAE encodes an input clip $x$ into a spatiotemporal latent $z_0 = \mathcal{E}(x)$, and the frozen T5-XXL encoder maps the prompt to a conditioning embedding $c$ of maximum length $300$. STDiT is trained with a rectified-flow objective, in which the noisy latent $z_\tau = (1-\tau)\, z_0 + \tau\, \epsilon$ with $\epsilon \sim \mathcal{N}(0, I)$ linearly interpolates between data and noise, and STDiT regresses the velocity $z_0 - \epsilon$:
\begin{equation}
\mathcal{L}(\theta) = \mathbb{E}_{\tau,\, z_0,\, \epsilon}\,\big\lVert\, v_\theta(z_\tau, \tau, c) - (z_0 - \epsilon)\, \big\rVert_2^2 .
\label{eq:rfloss}
\end{equation}
The timestep $\tau$ is drawn from a logit-normal distribution and shifted toward higher noise levels for higher-resolution and longer clips, following the resolution- and length-aware timestep sampling of Open-Sora~\cite{ref47_zheng2024opensora,opensora_v12_report}.

\subsection{Dynamic Bucket-Based Sampling}
\begin{figure}[htbp]
\centering
\includegraphics[width=0.85\textwidth]{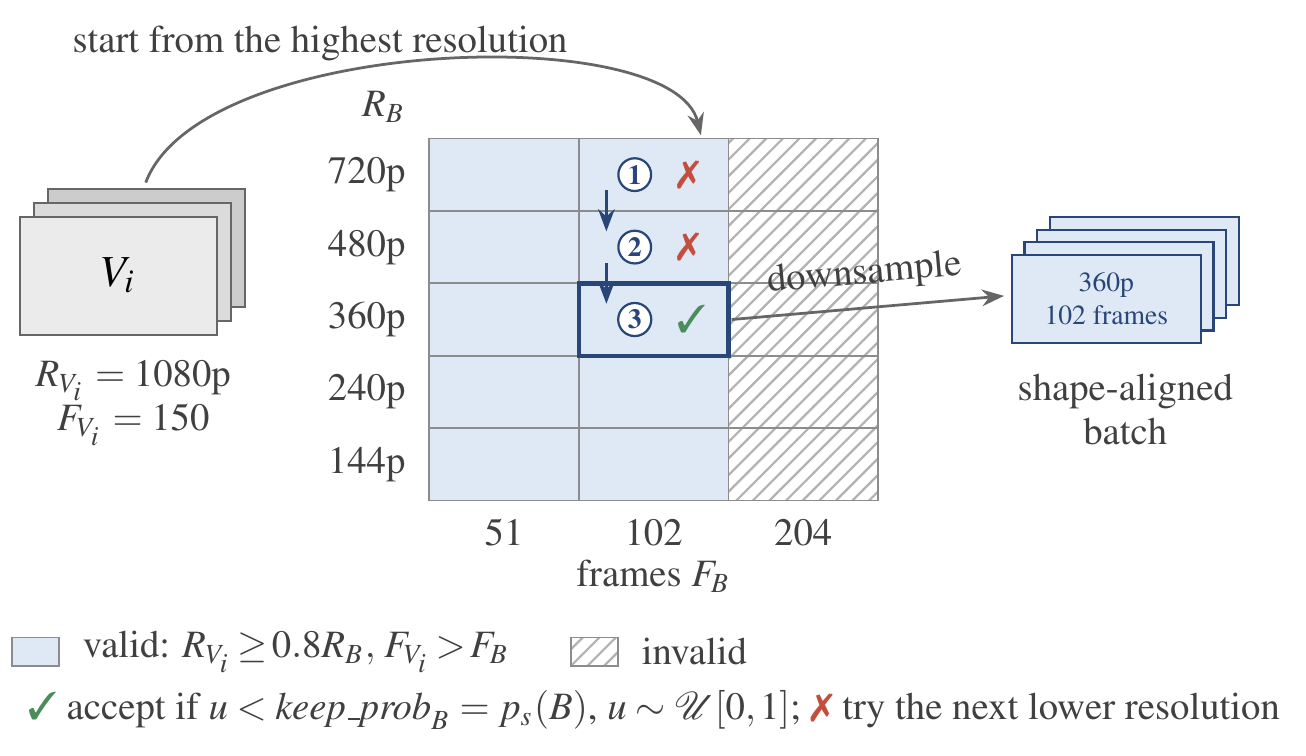}
\caption{Dynamic bucket sampling over the resolution--frame lattice. A clip $V_i$ is routed to the highest-resolution valid bucket and accepted with probability $\mathit{keep\_prob}_B$; otherwise the next lower resolution is tried. The accepted clip is downsampled into a shape-aligned batch. Per-stage $\mathit{keep\_prob}$ (Fig.~\ref{fig:sampling_heatmap}) acts as the curriculum control. Only a subset of buckets is shown. Based on the bucket mechanism of~\cite{ref47_zheng2024opensora}.}
\label{fig:bucket_sampling}
\end{figure}

Surgical clips vary substantially in resolution, aspect ratio, and duration. Rather than padding or cropping them to a fixed shape, we adopt the bucket-based sampling natively supported by Open-Sora~\cite{ref47_zheng2024opensora} (Fig.~\ref{fig:bucket_sampling}) and repurpose its per-bucket selection probabilities as a stage-conditioned curriculum control. A bucket is defined as a triplet $B = (R, F, A)$, where $R$ is the resolution in pixel area, $F$ is the number of frames, and $A$ is the aspect ratio, and each bucket is assigned a keep probability $\mathit{keep\_prob}_B$ and a batch size. For a video $V_i$, the set of valid buckets is
\begin{equation}
B_{\mathrm{valid}} = \{B \mid R_{V_i} \ge 0.8\, R_B \;\wedge\; F_{V_i} > F_B\}.
\label{eq:bvalid}
\end{equation}
Valid buckets are visited in descending order of resolution, and at each resolution $R$ the bucket with the largest valid frame count is selected,
\begin{equation}
\mathrm{Select}(V_i \mid R) = \arg\max_{B \in B_{\mathrm{valid}},\; R_B = R} F_B ,
\label{eq:bucket_select}
\end{equation}
and accepted if $u < \mathit{keep\_prob}_B$ with $u \sim \mathcal{U}[0,1]$. If it is rejected, the next lower resolution is tried, and a clip for which no bucket is accepted is skipped. The accepted clip is then resized and temporally subsampled to the bucket's shape, so that every batch contains clips of identical shape.

This dynamic strategy aligns batch tensor shapes to minimize padding, controls memory and compute by lowering $\mathit{keep\_prob}$ for high-resolution/long-frame buckets, and accelerates convergence by oversampling low-resolution/short-frame buckets early on.

\begin{figure}[htbp]
\centering
\includegraphics[width=0.75\textwidth]{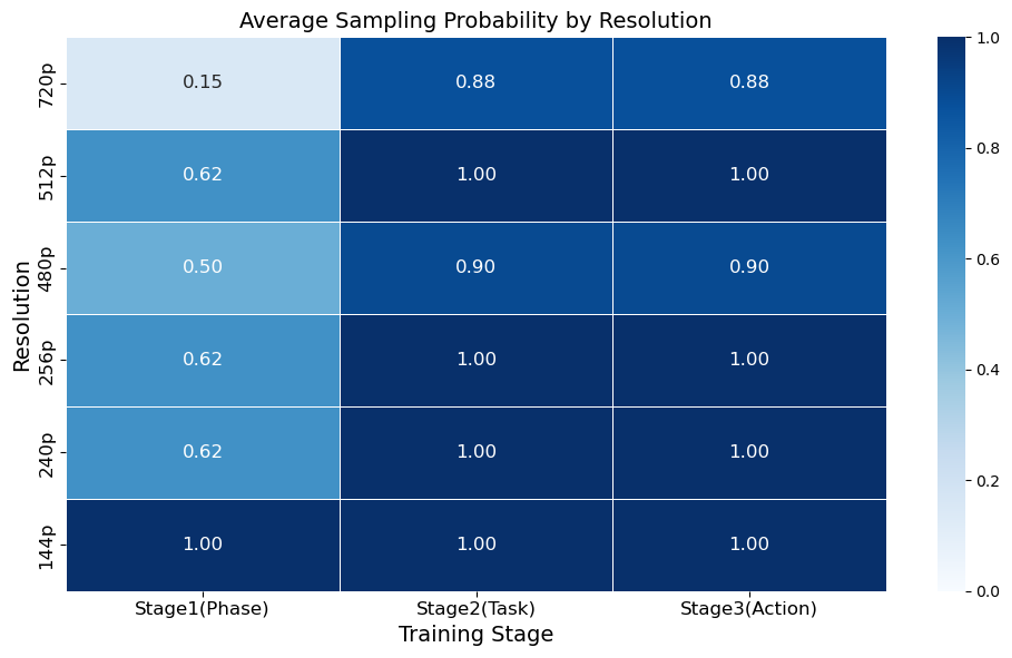}
\caption{Average bucket keep probability by resolution across the three curriculum stages. Stage~1 (Phase) heavily downsamples high-resolution buckets to emphasize global context, while Stages~2 (Task) and~3 (Action) retain high-resolution clips to capture fine visual detail.}
\label{fig:sampling_heatmap}
\end{figure}

The per-stage $\mathit{keep\_prob}$ serves as the curriculum control (Fig.~\ref{fig:sampling_heatmap}): Stage~1 downweights high-resolution buckets to acquire global context, while Stages~2 and~3 restore them to sharpen fine detail.

\subsection{Curriculum Transfer Learning Framework}
\begin{figure}[htbp]
\centering
\includegraphics[width=\textwidth]{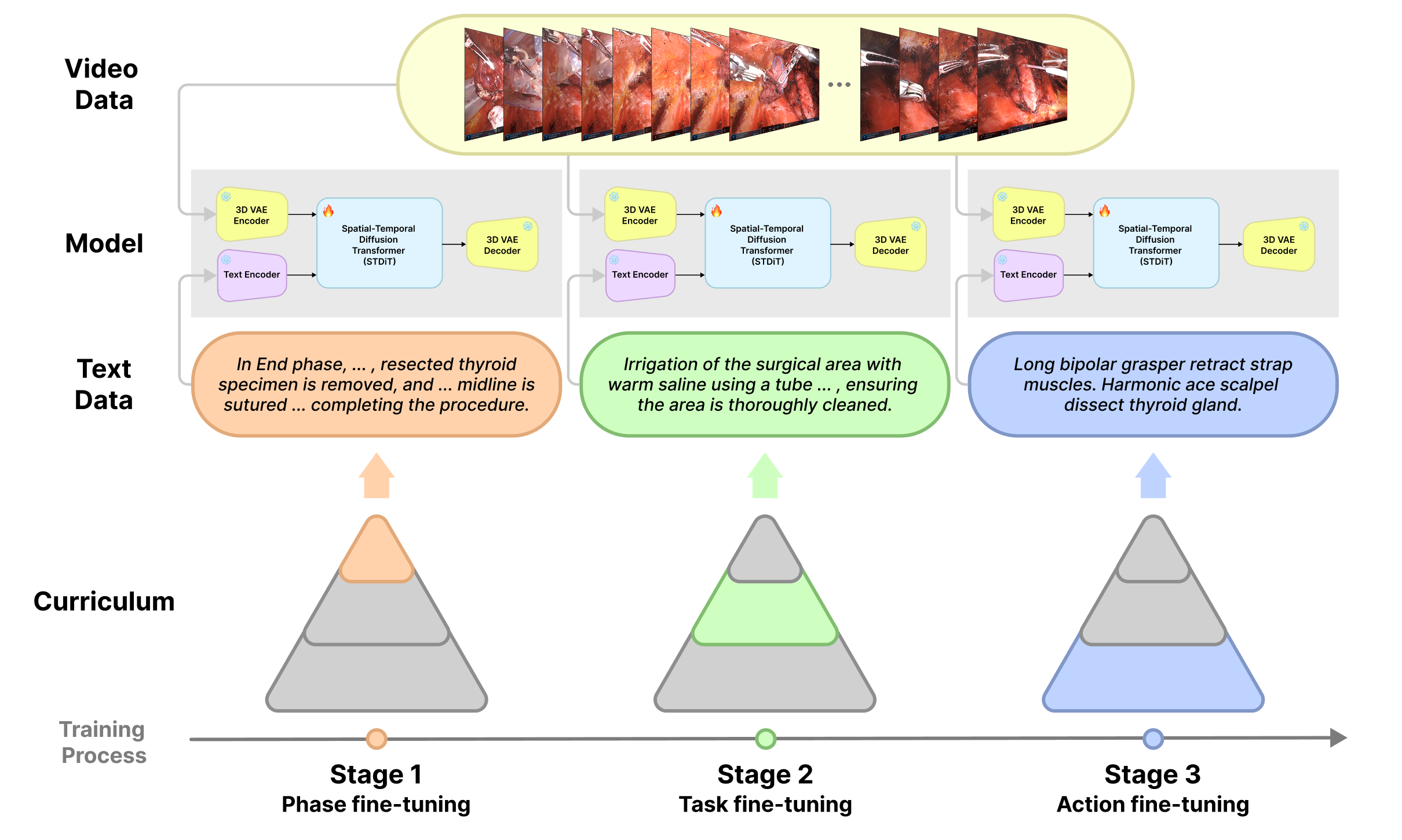}
\caption{Three-stage curriculum transfer learning pipeline. The text encoder and 3D VAE are frozen, and STDiT is fine-tuned progressively from Phase to Task to Action levels.}
\label{fig:curriculum_framework}
\end{figure}

Rather than fine-tuning a generic video diffusion model directly on fast, complex Action-level clips, we adopt a three-stage curriculum (Fig.~\ref{fig:curriculum_framework}) that mirrors the surgical hierarchy and proceeds from coarse to fine-grained representations. Throughout all stages, the T5-XXL text encoder and 3D VAE are frozen, and only STDiT is updated, focusing learning capacity on text-conditioned spatiotemporal generation while reusing the pretrained vision--language priors.

\textit{Stage 1: Phase fine-tuning.} The pretrained STDiT is first fine-tuned on Phase-level data, which contains long, slow clips capturing global surgical context. We set high $\mathit{keep\_prob}$ on low-resolution buckets and low $\mathit{keep\_prob}$ on high-resolution buckets, and run the longest schedule. This stage aligns the model with the general visual distribution of the surgical cavity, lighting, and tissue appearance.

\textit{Stage 2: Task fine-tuning.} Starting from Stage~1 weights, the model is fine-tuned on Task-level data to learn intermediate procedural goals and coordinated instrument motions. Compared with Stage~1, the sampling strategy increases $\mathit{keep\_prob}$ on higher-resolution buckets to capture sharper visual structure.

\textit{Stage 3: Action fine-tuning.} Finally, the model is fine-tuned on the Action-level dataset, which contains highly specific, fast tool--tissue interactions. Because the global and intermediate-level context has already been internalized in the previous stages, even a small amount of well-curated data and a short schedule are sufficient to refine fine details and physical plausibility. The bucket sampling configuration follows Stage~2.

Formally, let $p_s(B)$ be the keep probability of bucket $B$ at curriculum stage $s \in \{\mathrm{Phase}, \mathrm{Task}, \mathrm{Action}\}$ (Fig.~\ref{fig:sampling_heatmap}). Together with the stage-specific clips, $p_s$ turns the bucket selection of Eqs.~\eqref{eq:bvalid}--\eqref{eq:bucket_select} into a stage-specific training distribution $\mathcal{D}_s$. Starting from the pretrained backbone $\theta_0$, each stage minimizes the rectified-flow loss of Eq.~\eqref{eq:rfloss} over its own distribution:
\begin{equation}
\theta_s = \arg\min_{\theta}\; \mathbb{E}_{(x,c)\sim \mathcal{D}_s}\big[\mathcal{L}(\theta)\big], \quad \text{initialized from } \theta_{s-1}.
\label{eq:curriculum}
\end{equation}
Parameters are optimized with AdamW ($\epsilon = 10^{-15}$, learning rate $10^{-4}$, gradient clip $1.0$, $1{,}000$ warmup steps) under bf16 precision with gradient checkpointing and ZeRO Stage~2, and an exponential moving average of the weights with decay $0.99$ is maintained.
Stage- and bucket-level data composition and time budgets are summarized in Table~\ref{tab:training_schedule}, and implementation details in Table~\ref{tab:impl}.

\begin{table}[htbp]
\caption{Dataset split and per-stage curriculum composition. ``Annotated'': cases labeled at each level; ``Train'': cases used after the 4-case test split. Clips, steps, and time refer to training.}
\label{tab:training_schedule}
\centering
\begin{tabular}{lccccc}
\toprule
\textbf{Level} & \textbf{Annotated} & \textbf{Train} & \textbf{Clips} & \textbf{Steps} & \textbf{Time (h)} \\
\midrule
Phase  & 20 & 20 & 16{,}000 & 55{,}000 & 1{,}375 \\
Task   & 16 & 12 & 3{,}200  & 15{,}000 & 375 \\
Action & 7  & 7  & 160      & 1{,}200  & 28.8 \\
\midrule
\multicolumn{6}{l}{\footnotesize Corpus: 43 cases (39 train / 4 test, patient-level). ``Annotated'': cases} \\
\multicolumn{6}{l}{\footnotesize labeled at each level; ``Train'': cases used after the test split.} \\
\bottomrule
\end{tabular}
\end{table}

\begin{table}[htbp]
\caption{Implementation details and training/evaluation environment. Optimization hyperparameters (learning rate, $\epsilon$, warmup, EMA decay) are given in the text.}
\label{tab:impl}
\centering
\begin{tabular}{ll}
\toprule
\textbf{Setting} & \textbf{Value} \\
\midrule
Backbone & Open-Sora v1.2 (STDiT3-XL/2) \\
Scheduler & rectified flow (rflow) \\
Batch size & 1 per GPU \\
Precision & bf16 \\
Distributed & DeepSpeed ZeRO Stage 2 \\
Gradient checkpointing & yes \\
GPU & $8 \times$ NVIDIA A100 80\,GB \\
CPU & AMD EPYC 7402X @ 2.8\,GHz \\
Framework & PyTorch 2.1.0, Open-Sora 1.2.0 \\
OS / CUDA & Ubuntu 20.04.6, CUDA 11.8 \\
\bottomrule
\end{tabular}
\end{table}

\section{Experiments and Results}
\label{sec:experiments}

\subsection{Experimental Setup}
We evaluated the proposed framework on a patient-disjoint test set comprising 4 of the 43 BABA robotic thyroidectomy videos described in Section~\ref{sec:methodology}. From each test case, 10 representative task sequences were selected and clipped to 5-second segments at 24~FPS (120 frames), yielding 40 ground-truth clips. For each case, the model generated 10 End-Phase clips at matched specifications. Implementation details and the training/evaluation environment are summarized in Table~\ref{tab:impl}.

For quantitative evaluation, we report Peak Signal-to-Noise Ratio (PSNR), Structural Similarity Index Measure (SSIM)~\cite{ref70_wang2004image}, Learned Perceptual Image Patch Similarity (LPIPS)~\cite{ref71_zhang2018unreasonable}, FloLPIPS~\cite{ref72_danier2022flolpips}, Fr\'echet Video Distance (FVD)~\cite{ref73_unterthiner2019fvd}, and Fr\'echet Inception Distance (FID). Surgical video generation is inherently non-deterministic: from a given prompt the model synthesizes a plausible realization rather than a pixel-aligned copy of any particular ground-truth clip. Reference-based pixel and patch metrics (PSNR, SSIM, LPIPS, FloLPIPS) therefore penalize legitimate appearance variation and, while reported for completeness, serve only as secondary frame-level indicators. We accordingly treat the distribution-level metrics FID and FVD, together with the clinical user study, as our primary measures, as they assess whether generated frames and videos match the real surgical distribution rather than a single reference clip.

For qualitative clinical evaluation, six clinicians (one resident and five fellows) rated the generated videos using a 5-point Likert scale~\cite{ref69_likert1932technique} on visual similarity, anatomical accuracy, procedural consistency of motion, and overall quality. Inspired by the evaluation dimensions of VBench~\cite{ref58_huang2024vbench}, the questionnaire comprised seven items spanning general video quality (visual similarity, motion smoothness, imaging quality, aesthetic quality) and clinical relevance (human anatomy, clinical applicability, clinical reliability).

\subsection{Quantitative Evaluation of Curriculum Learning}
Table~\ref{tab:quantitative_eval} reports per-stage performance, focusing on the Stage~2 (Task)~$\rightarrow$~Stage~3 (Action) transition. On our primary distribution-level metrics, Stage~3 reduces FID by 15.7\% and FVD by 20.1\% relative to the final Stage~2 model, indicating improved spatial and spatiotemporal realism. The secondary frame-level metrics move in the same direction but by smaller margins (PSNR $+5.6\%$, SSIM $+1.2\%$, LPIPS $-1.5\%$); FloLPIPS alone worsens slightly ($+1.0\%$), which we attribute to its sensitivity to frame-to-frame flow differences under non-deterministic generation rather than to degraded temporal quality, a reading corroborated by the large FVD improvement on the same clips.

\begin{table}[htbp]
\caption{Per-stage quantitative evaluation of the curriculum transfer learning framework.}
\label{tab:quantitative_eval}
\centering
\resizebox{\textwidth}{!}{%
\begin{tabular}{llcccccc}
\toprule
\textbf{Model Stage} & \textbf{Steps} & \textbf{SSIM $\uparrow$} & \textbf{PSNR $\uparrow$} & \textbf{LPIPS $\downarrow$} & \textbf{FloLPIPS $\downarrow$} & \textbf{FVD $\downarrow$} & \textbf{FID $\downarrow$} \\
\midrule
\multirow{4}{*}{Stage 2 (Task)}
 & 3k    & \textbf{0.307} & 9.77  & 0.717 & 0.726 & 4242.5 & 244.9 \\
 & 7k    & 0.285 & 9.64  & 0.744 & 0.739 & 4652.7 & 249.1 \\
 & 11k   & 0.264 & 10.18 & 0.680 & \textbf{0.709} & 5097.4 & 213.5 \\
 & 15k   & 0.285 & 10.19 & 0.719 & 0.722 & 4105.4 & 241.2 \\
\midrule
\multirow{3}{*}{Stage 3 (Action)}
 & 0.4k  & 0.299 & 10.41 & 0.708 & 0.734 & 4665.6 & 233.6 \\
 & 0.8k  & 0.282 & 10.53 & \textbf{0.677} & 0.739 & 4608.1 & \textbf{198.5} \\
 & 1.2k  & 0.288 & \textbf{10.77} & 0.708 & 0.729 & \textbf{3278.4} & 203.4 \\
\bottomrule
\end{tabular}%
}
\end{table}

Within Stage~2, FVD plateaued near 4100 and FID remained above 213, indicating that Task-only training had limited headroom. The frame-level scores fluctuated non-monotonically: SSIM dipped to 0.264 at 11k before recovering to 0.285 at 15k. We attribute this to the model generating sharper, more varied textures as training progresses; because such textures cannot match the ground-truth clip pixel by pixel, they can lower reference-based scores even when visual quality does not decrease. In Stage~3, the primary metrics improved overall: after an initial rise (4665.6 at 0.4k), FVD decreased to 3278.4 at 1.2k (a 29.7\% reduction from the first Stage~3 checkpoint), and FID reached its minimum of 198.5 at 0.8k. These results indicate that Action-level fine-tuning improves fine-grained detail and temporal coherence beyond the Stage~2 plateau.

\subsection{Ablation Study}
To verify that Stage~3's gains stem from the curriculum rather than from additional training, we compared the full model (Stage~$1\rightarrow 2\rightarrow 3$) against a matched-budget baseline that spends the final 1.2k steps on Stage~2 data (Stage~$1\rightarrow 2\rightarrow 2$). Under an identical compute budget (Table~\ref{tab:ablation}), the curriculum model improves FVD by 18.0\% alongside concurrent PSNR ($+9.5\%$) and SSIM ($+9.7\%$) gains, while LPIPS, FloLPIPS, and FID remain essentially unchanged. The simultaneous improvement in distribution-level (FVD) and pixel/structural (PSNR, SSIM) fidelity suggests that the improvement is not specific to a single metric. Notably, Stage~3 attains these gains from only 160 Action-level clips, 20 times fewer than the 3{,}200 Task-level clips used by the baseline, suggesting that, in this setting, curriculum ordering contributes more than additional Task-level data.

\begin{table}[htbp]
\caption{Ablation study on the effect of curriculum transfer learning. Both models share Stage~1 (55k) and Stage~2 (15k) and differ only in the final 1.2k steps.}
\label{tab:ablation}
\centering
\resizebox{\textwidth}{!}{%
\begin{tabular}{lcccccc}
\toprule
\textbf{Schedule} & \textbf{SSIM $\uparrow$} & \textbf{PSNR $\uparrow$} & \textbf{LPIPS $\downarrow$} & \textbf{FloLPIPS $\downarrow$} & \textbf{FVD $\downarrow$} & \textbf{FID $\downarrow$} \\
\midrule
Stage $1\!\rightarrow\!2\!\rightarrow\!2$ (matched budget) & 0.263 & 9.83 & \textbf{0.696} & \textbf{0.721} & 3997.2 & \textbf{200.9} \\
Stage $1\!\rightarrow\!2\!\rightarrow\!3$ (Ours) & \textbf{0.288} & \textbf{10.77} & 0.708 & 0.729 & \textbf{3278.4} & 203.4 \\
\bottomrule
\end{tabular}%
}
\end{table}

\subsection{Qualitative Evaluation}
\begin{figure}[htbp]
\centering
\includegraphics[width=0.85\textwidth]{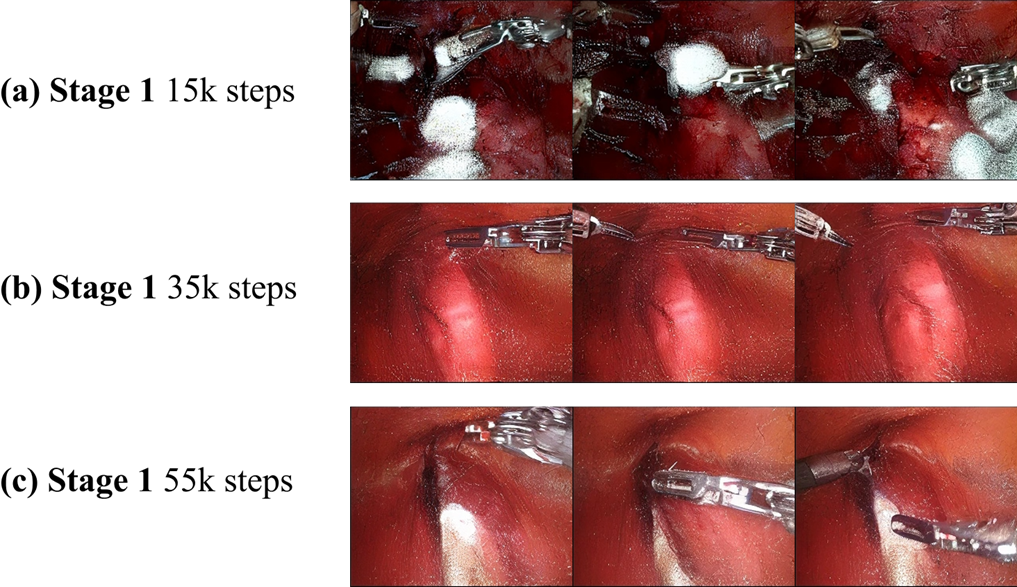}
\caption{Qualitative results across Stage~1 training steps (15k, 35k, 55k). Physical inconsistencies (e.g., extra robotic arms) progressively disappear and tissue textures become realistic.}
\label{fig:stage1_qualitative}
\end{figure}

Visual inspection complements the quantitative findings. In early Stage~1 (15k steps), the model often produced physically impossible scenes such as three robotic arms appearing simultaneously, indicating that the basic constraints of the surgical environment had not yet been learned. By 35k steps, the arm count and motion errors were largely corrected, but textures of gauze and surrounding tissue remained blurred. By 55k steps, both physical constraints and fine textures (e.g., gauze fabric pattern, tissue surface) were faithfully reproduced (Fig.~\ref{fig:stage1_qualitative}), suggesting that Phase-level training captures global surgical context.

\begin{figure}[htbp]
\centering
\includegraphics[width=0.85\textwidth]{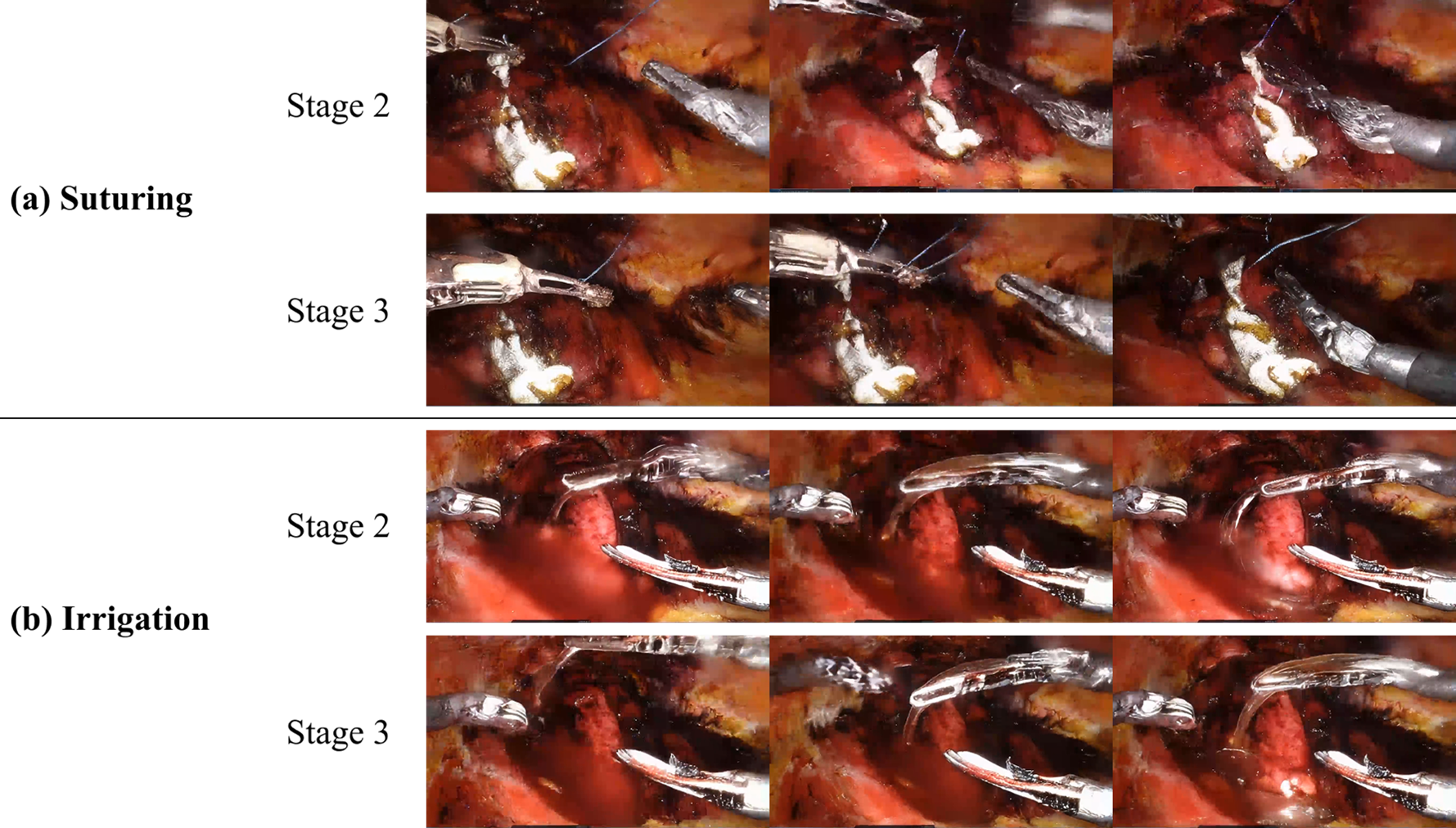}
\caption{Qualitative comparison between Stage~2 and Stage~3 on \textit{suturing} and \textit{irrigation} tasks. Stage~3 reproduces fine instruments (thread, needle) and fluid dynamics more faithfully.}
\label{fig:stage23_qualitative}
\end{figure}

We also compared Stage~2 and Stage~3 models on specific Task prompts (Fig.~\ref{fig:stage23_qualitative}). For \textit{suturing}, Stage~2 captured rough shapes of needle and thread but rendered them ambiguously across frames, whereas Stage~3 produced sharper, more consistent thread contours and a clearer grasp on needle and thread. For \textit{irrigation}, Stage~2 portrayed liquid sloshing and suction but with unnatural dynamics, while Stage~3 reproduced the inflow of liquid and surface specular reflections more realistically. These qualitative differences echo the quantitative gains in FVD and FID for Stage~3.

\subsection{Clinical User Study}
\begin{figure}[htbp]
\centering
\includegraphics[width=0.75\textwidth]{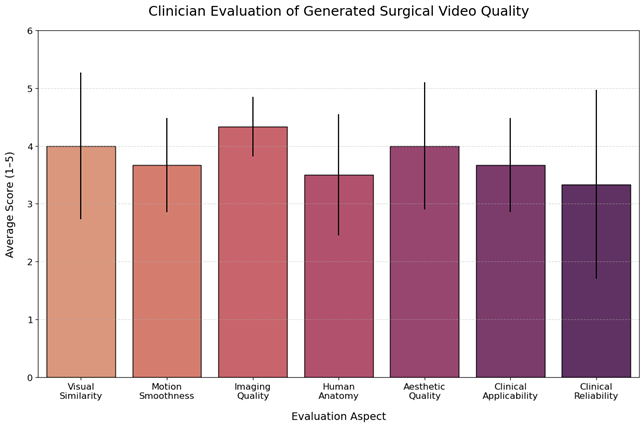}
\caption{User study results from six clinicians. The model performs best on imaging-related criteria but is more limited on anatomical accuracy and clinical reliability.}
\label{fig:user_study}
\end{figure}

Six clinicians rated the generated videos on the seven items described above. The mean scores (Fig.~\ref{fig:user_study}) are highest for \textit{Imaging Quality} (4.33), followed by \textit{Visual Similarity} (4.00) and \textit{Aesthetic Quality} (4.00), suggesting that the model reproduces the overall color, texture, and sharpness of real surgical video. \textit{Motion Smoothness} (3.67) and \textit{Clinical Applicability} (3.67) also received favorable ratings.

In contrast, \textit{Clinical Reliability} (3.33) and \textit{Human Anatomy} (3.50) received lower scores with larger inter-rater variance, suggesting that while overall shapes and motions are convincing, precise anatomical structures and subtle tissue interactions are not yet at clinical-grade fidelity. These findings indicate that further refinement of soft-tissue physics and anatomical priors will be needed for safe deployment in clinical training scenarios.

\section{Discussion}
\label{sec:discussion}

In addition to quantitative metrics, we incorporate a structured user study with practicing clinicians, enabling assessment of clinical reliability and anatomical accuracy beyond raw video quality. More importantly, our results show that a hierarchically structured curriculum extends high-fidelity generation to BABA robotic thyroidectomy without any curated benchmark, indicating that the approach is not confined to data-rich procedures targeted by prior work. By integrating hierarchical annotation with curriculum transfer learning, the framework treats surgery as a structured, hierarchical procedural domain rather than a flat sequence of frames, providing a path toward domain-specialized generative models with clinical relevance.

The quantitative results support this view: Action-level fine-tuning breaks the Stage~2 plateau, and the ablation (Table~\ref{tab:ablation}) shows the gain stems from the curriculum rather than extra training, as the matched-budget Stage~$1\rightarrow 2\rightarrow 2$ baseline lags despite using $20\times$ more clips. 

Qualitatively, the progression of Stage~1 from physically implausible scenes (15k steps) to faithful textures (55k steps) suggests that Phase-level fine-tuning provides a foundation upon which Stage~2 and Stage~3 build with comparatively few additional examples. The Stage~2-vs-Stage~3 comparisons on \textit{suturing} and \textit{irrigation} suggest that Action-level training particularly benefits fine instruments and fluid dynamics, where temporal coherence is critical.

The user study makes the remaining limitations concrete: \textit{Imaging Quality}, \textit{Visual Similarity}, and \textit{Aesthetic Quality} all exceed 4.0/5.0, while \textit{Clinical Reliability} and \textit{Human Anatomy} remain below this threshold with substantial inter-rater variance. This indicates that improving anatomical fidelity and soft-tissue dynamics, rather than overall image quality, is the main remaining challenge.

\subsection{Limitations and Future Work}
This study has several limitations. First, the dataset originates from a single institution (Seoul National University Hospital) with 43 BABA robotic thyroidectomy cases, so generalization to other institutions, surgical techniques, and patient populations is not directly demonstrated. Second, we focus on the End Phase for Task and Action annotation; modeling the entire procedure with consistent transitions between phases remains future work. Third, the absence of a public benchmark for surgical video generation prevents direct quantitative comparison with prior models; our evaluation is therefore relative to internal baselines. Fourth, while we employ standard quality metrics and expert review, we do not yet measure how generated videos affect actual training outcomes. Future work will integrate the system into existing training pipelines and quantify educational impact on residents and fellows.

To address these gaps, we plan to (i) construct an internal benchmark grounded in prior evaluation protocols to enable more objective comparison, (ii) expand the dataset to multiple institutions, surgical techniques, and patient cohorts, (iii) extend the framework to generate full-procedure surgical videos with consistent inter-phase transitions, and (iv) condition the generative model on additional patient-specific signals such as preoperative CT/MRI or surgical kinematics, enabling personalized rehearsal aligned with each patient's anatomy.

\section{Conclusion}
\label{sec:conclusion}
We presented a framework for surgical video generation that combines a Phase--Task--Action hierarchical annotation system with a three-stage curriculum transfer learning strategy on top of the Open-Sora STDiT backbone, applied to BABA robotic thyroidectomy. Quantitative metrics, qualitative analysis, an ablation study, and a clinician user study collectively show that the framework adapts a large pretrained model to domain-specific surgical content and improves fine-grained visual fidelity while preserving global procedural context. Although validated on a single procedure, the approach may extend to other procedures and offers a basis for AI-driven surgical simulation, education content generation, and downstream surgical AI systems requiring high-fidelity synthetic video.

\subsubsection*{Acknowledgments.}
The authors thank the clinicians of the breast and endocrine surgery division at Seoul National University Hospital for their guidance on annotation and clinical evaluation. This research was supported by the Naver Digital Bio Research Fund (Grant No.~3720250030).

\bibliographystyle{splncs04}
\bibliography{references}
\end{document}